\documentclass[conference]{IEEEtran}
\IEEEoverridecommandlockouts

\makeatletter
\newcommand{\linebreakand}{%
  \end{@IEEEauthorhalign}
  \hfill\mbox{}\par
  \mbox{}\hfill\begin{@IEEEauthorhalign}
}
\makeatother

\usepackage{booktabs}
\usepackage{cite}
\usepackage{amsmath,amssymb,amsfonts}
\usepackage{algorithmic}
\usepackage{graphicx}
\usepackage{textcomp}
\usepackage{xcolor}
\def\BibTeX{{\rm B\kern-.05em{\sc i\kern-.025em b}\kern-.08em
    T\kern-.1667em\lower.7ex\hbox{E}\kern-.125emX}}

\usepackage{url}
\IEEEoverridecommandlockouts

\begin{document}

\title{Design and Evaluation of an AI--Enabled Cloud--Edge Architecture for Connected Precision Agriculture Farms}

\author{
\IEEEauthorblockN{
Deckshitha Angadi\textsuperscript{1,2},
Koteshwar Goud Surga\textsuperscript{3},
Nagaraju Lakkaraju\textsuperscript{1},
Naveena Budda\textsuperscript{4},\\
Vikas Agarwal\textsuperscript{5,6},
Giridhara Venkata Ram Raj Mulasa\textsuperscript{7},
Ravi Killamsetty\textsuperscript{1},\\
Chandrasekhara Sarma Mallubhotla\textsuperscript{8},
Narsimlu Kemsaram\textsuperscript{9}
}
\IEEEauthorblockA{
\textsuperscript{1}\textit{AI, IoT and Robotics Lab (AIR Lab), UAVs Group, Autonomous Robotics Systems Limited, Hyderabad, India}\\
\textsuperscript{2}\textit{Department of Microelectronics and VLSI Design, University of Hyderabad, Hyderabad, India}\\
\textsuperscript{3}\textit{Department of Artificial Intelligence, ThoughtGreen Technologies Private Limited, Hyderabad, India}\\
\textsuperscript{4}\textit{Department of Internet of Things, Ideabytes Software India Private Limited, Hyderabad, India}\\
\textsuperscript{5}\textit{School of Computer Science, Georgia Institute of Technology, Atlanta, GA, USA}\\
\textsuperscript{6}\textit{Enterprise Solutions Unit, Tata Consultancy Services, Atlanta, GA, USA}\\
\textsuperscript{7}\textit{Business Analysis Product Division Management, EPAM India, Hyderabad, India}\\
\textsuperscript{8}\textit{Department of Genetics and Plant Breeding, Kaveri University, Gowraram, India}\\
\textsuperscript{9}\textit{Department of Artificial Intelligence, University of Malaya, Kuala Lumpur, Malaysia}
}
}

\maketitle

\begin{abstract}

Plant diseases cause significant yield losses worldwide, with tomato crops particularly susceptible to early blight, late blight, and leaf mold. 
Manual monitoring is practical only for small-scale farms and becomes unmanageable at larger scales. 
To tackle this limitation, an artificial intelligence (AI) enabled cloud-edge architecture is proposed for autonomous crop monitoring. 
This proposed architecture integrates internet of things (IoT) sensors, unmanned aerial vehicles (UAVs), deep learning, Azure IoT Hub-based cloud analytics, and multi-platform (mobile app, web app, and embedded edge device platform) interfaces to enable real-time detection of tomato diseases. 
For training and validation, we used publicly available datasets, such as PlantVillage and Kaggle.
A TensorFlow model trained on a collected dataset is deployed across mobile, web, and edge-device platforms.
Experimental results show detection effectiveness around 92-95\%, with consistent performance over diverse environments and device platforms. 
The proposed system improves disease detection effectiveness, lowers dependence on manual inspection, and enables prompt interventions, thereby supporting sustainable, connected precision agriculture farms. 

\end{abstract}


\section{Introduction}


Autonomous sensing platforms \cite{kim2019unmanned}, \cite{kemsaram2017design} are transforming precision agriculture by enabling large-scale, real-time crop monitoring \cite{liu2021boost}, \cite{rs10091423}. 
Traditional approaches to disease detection rely on manual inspection, which is feasible only on small farms and remains highly labor-intensive, error-prone, and time-consuming \cite{HARAKANNANAVAR2022305}, \cite{mohyuddin2024evaluation}. 
For large-scale tomato crop cultivation, where diseases such as early blight, late blight, and bacterial spot can spread rapidly, autonomous crop monitoring is required for timely detection and intervention \cite{sundararaman2023transformative}, \cite{Si-2026-hortsci-61.1}.

Recent advances in deep learning, combined with vision-based algorithm \cite{das2023stixel}, \cite{kemsaram2017multi}, provide new opportunities for scalable disease diagnosis \cite{rs15092450}, \cite{ale2019deep}. 
Deep neural networks (DNNs) have achieved state-of-the-art accuracy in plant disease detection, and their deployment on mobile, web, and edge computing device platforms has enabled the development of real-time field-ready applications. 
Complementing deep learning \cite{hasan2020review}, edge computing devices \cite{kemsaram2016autonomous}, IoT-enabled sensors \cite{sharafat2025iot}, and Azure IoT Hub-based cloud services \cite{11291228}, can capture environmental data, enabling a comprehensive understanding of crop health.

\begin{figure*}[!t]
\centering
\includegraphics[width=0.68\textwidth]{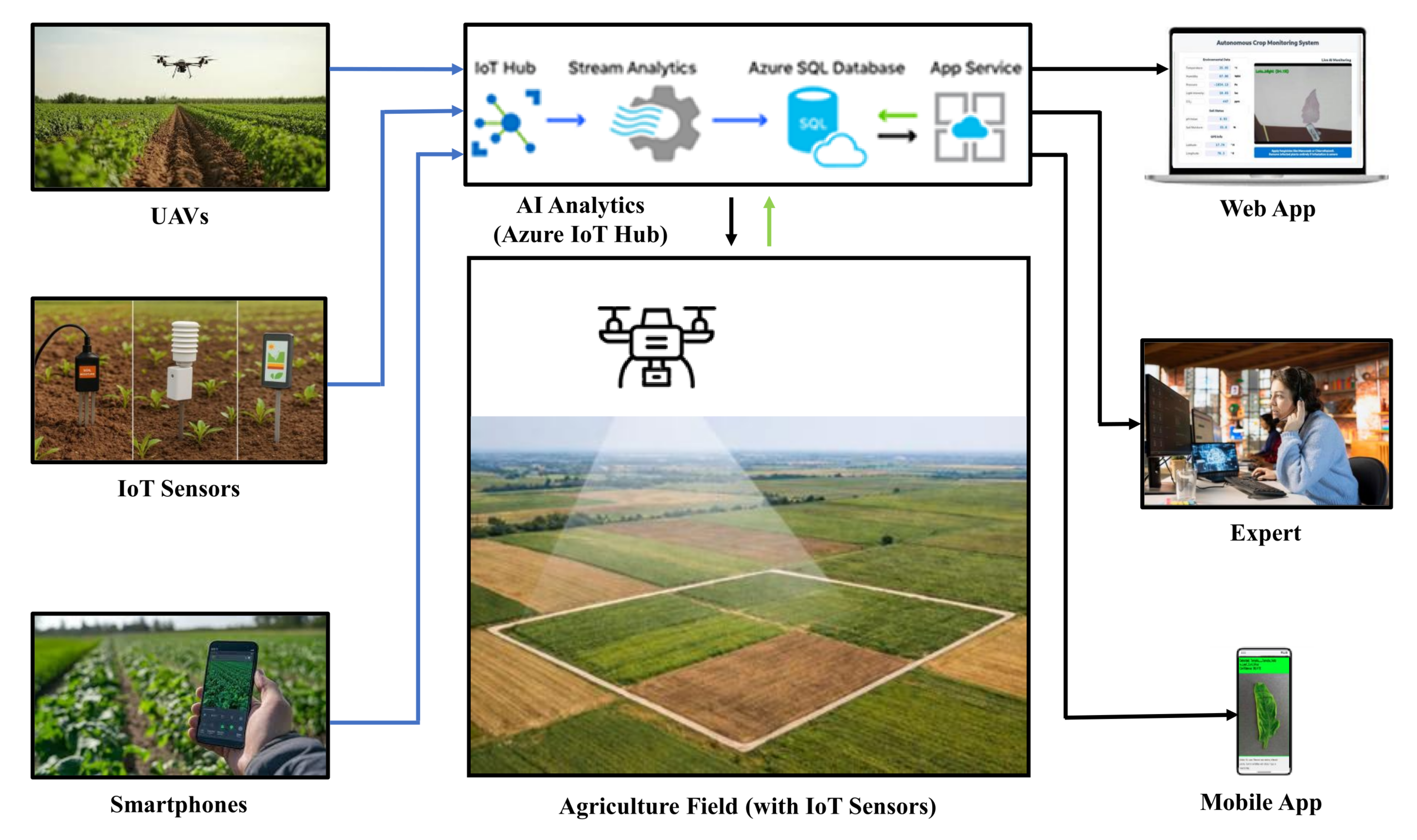}
\caption{
Block diagram of the proposed AI-enabled cloud-edge architecture for an autonomous crop monitoring system, integrating IoT sensors, UAV-based imaging, smartphones, edge devices, and cloud-based analytics for real-time tomato crop disease detection and decision support.
}
\label{ACMS_BlockDiagram}
\end{figure*}

This paper proposes an AI-enabled cloud-edge architecture that employs cameras for image acquisition, trained model for disease detection, web and mobile applications for user interaction. 
In addition, it integrates IoT sensors and Azure IoT Hub to enable AI-driven cloud-based analytics for real-time detection of tomato crop diseases and decision support.
As illustrated in Figure \ref{ACMS_BlockDiagram}, images and sensor data from the agricultural field are collected through ground-based IoT sensors, UAV-mounted cameras, smartphones, processed in edge devices (Raspberry Pi 5), trained on a TensorFlow-based model for real-time tomato crop disease detection, and transmitted to the cloud for advanced analytics using Azure IoT Hub. 
The results are made available to farmers and agronomists through mobile, web, and edge device applications, enabling real-time disease classification and expert-recommended disease management solutions.

The main contributions of this paper are as follows: i) A vision–enabled sensing architecture for autonomous image and environmental data acquisition in tomato fields, ii) A deep learning–based disease detection pipeline trained and deployed across mobile, web, and edge platforms for real–time diagnosis, and iii) An Azure IoT Hub cloud integration framework that supports scalable analytics, farmer feedback, and timely intervention.
By combining AI–driven algorithms, IoT-based sensing, and cloud analytics, the proposed system enables early disease detection, reduces reliance on manual inspection, and supports sustainable, connected agricultural farms.

This paper is organized as follows. Section \ref{Sec_RelatedWork} reviews related work on crop monitoring and plant disease detection. 
Section \ref{Sec_ProposedSystem} introduces the proposed autonomous crop monitoring system and its key modules. 
Section \ref{Sec_Architecture} presents the cloud–edge architecture, highlighting the real-time inference and batch analytics paths. 
Section \ref{Sec_Design} describes the multi-stage design and end-to-end workflow from data acquisition to user interface applications. 
Section \ref{Sec_Development} details the implementation, including the software stack, platform components, and module integration. 
Section \ref{Sec_ExperimentsAndResults} reports the experimental setup and discusses the evaluation results. 
Finally, Section \ref{Sec_Conclusions} concludes the paper and provides research directions for future work.

\section{Related Work} \label{Sec_RelatedWork}

In this section, we first review related work on vision-based monitoring in agriculture. We then examine deep learning methods for plant disease detection. Finally, we discuss integration with web, mobile, edge, and cloud platforms.

\subsection{Vision-Based Monitoring in Agriculture}

UAVs are widely used in agriculture for their ability to capture images over large areas with minimal human effort \cite{velusamy2021unmanned}.
UAV-mounted RGB \cite{kemsaram2014experimental} and multispectral cameras support tasks such as crop stress detection, yield estimation, and disease classification \cite{barbedo2019review}. 
These methods highlight the scalability of UAVs in reducing labor for field inspections \cite{imran2025technological}.
However, many UAV-based systems focus on image collection, with limited real-time analytics for disease classification.

\subsection{Deep Learning for Plant Disease Detection}

Deep learning automates plant disease classification \cite{li2021plant}.
These models, trained on datasets such as PlantVillage and Kaggle, achieve over 90\% accuracy in identifying plant diseases \cite{tugrul2022convolutional}.
ResNet, Inception, and MobileNet excel at extracting image features \cite{rybczak2024deep}.
Mobile models such as TensorFlow Lite on smartphones now enable field diagnosis \cite{hegde2025leveraging}.
However, most research still uses controlled datasets or single-platform apps and does not address multi-platform scaling.

\subsection{Apps, Edge and Cloud Integration}

Web, mobile, edge, and cloud computing have advanced AI model deployment in agriculture by enabling low-latency, resource-efficient inference \cite{zhang2020overview}. 
For example, web and mobile applications \cite{naveed2025agrisage} and Raspberry Pi 5 edge devices \cite{rakesh2025implementation} running quantized DNN models demonstrate real-time crop disease classification.
These edge and app solutions integrate with cloud-based platforms \cite{bansal2020designing}, such as Microsoft Azure IoT and Google Cloud, which support large-scale data aggregation, adaptive model retraining, and decision support systems.

Although effective, these approaches are typically evaluated in isolation and not integrated into farmer tools. 
The proposed AI-enabled cloud-edge architecture bridges these gaps by integrating image data, IoT sensors, AI models, Azure IoT Hub, and cross-platform access to enable real-time, autonomous crop disease monitoring.

\section{Proposed Autonomous Crop Monitoring System} \label{Sec_ProposedSystem}



\begin{figure*}[!t]
\centering
\includegraphics[width=0.85\textwidth]{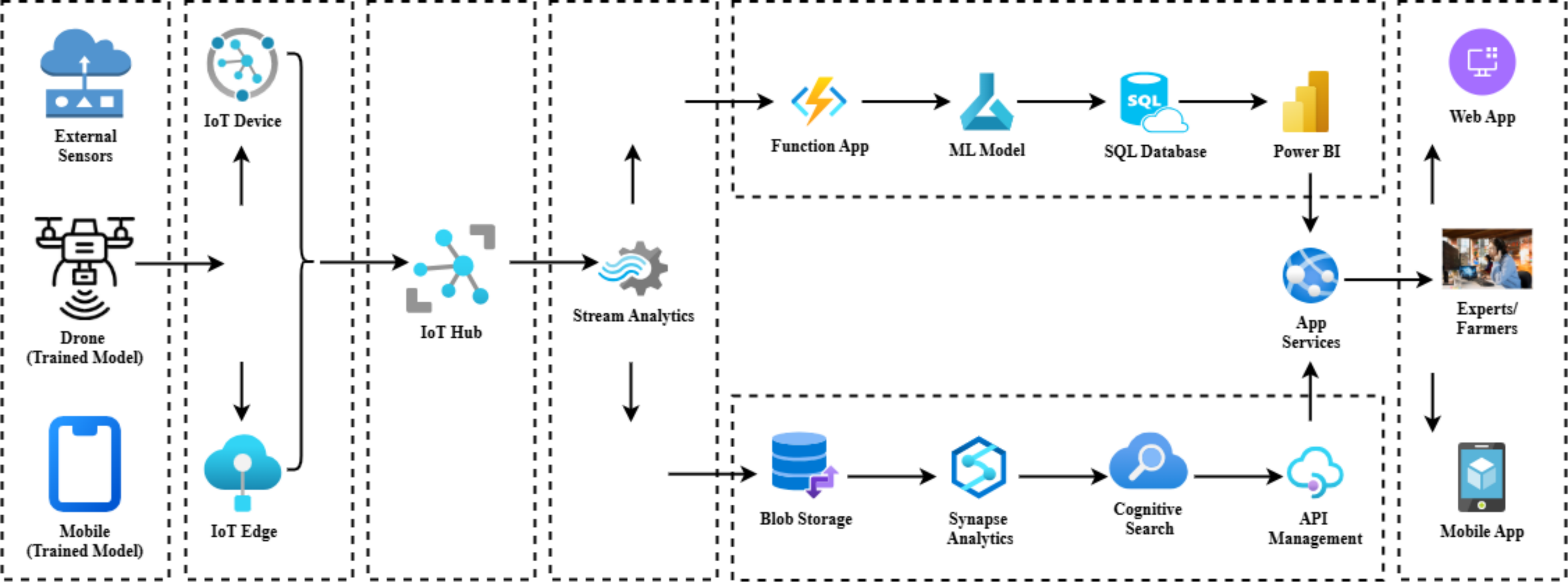}
\caption{
Architecture diagram of the proposed AI-enabled cloud-edge architecture for an autonomous crop monitoring system, integrating IoT sensors, UAV-based imaging, smartphones, edge devices, and cloud-based AI analytics for real-time tomato crop disease detection and decision support.
} \label{ACMS_ArchitectureDiagram}
\end{figure*}

The proposed AI-enabled autonomous crop monitoring system (ACMS) is designed and developed as an end-to-end framework for connected agriculture farms that integrates UAV-based imaging, IoT-enabled sensing, deep learning, edge computing, and cloud analytics to support real-time detection and monitoring of tomato diseases (see Figure \ref{ACMS_BlockDiagram} and \ref{ACMS_ArchitectureDiagram}).

\subsection{UAV-Based Image Acquisition}

UAVs equipped with high-resolution RGB cameras autonomously survey tomato fields, capturing large-scale visual data. 
These images provide early indications of plant stress and disease symptoms that may not be visible to the naked eye.
UAV flight planning ensures consistent coverage, while real-time communication enables data transfer to ground stations or cloud servers.

\subsection{IoT-Enabled Field Sensing}

To complement aerial imagery, ACMS integrates ground-based IoT sensors for soil moisture, pH, light, temperature, and humidity. 
These measurements provide contextual environmental data that enhance disease diagnosis by correlating visual symptoms with underlying field conditions.



\subsection{Edge Processing and Disease Detection}

Field data from UAVs and IoT sensors is transmitted to a Raspberry Pi–based edge computing unit. 
The Raspberry Pi 5 performs initial preprocessing tasks, including image resizing, filtering, and basic feature extraction. 
For rapid on-site diagnostics, a lightweight TensorFlow Lite model is deployed on the Raspberry Pi 5, enabling disease classification even in low-connectivity rural environments.

\subsection{Cloud Integration and Analytics}

For large-scale processing and storage, sensor and image data are uploaded to Azure IoT Hub in the cloud. 
Using Stream Analytics and an SQL database, the system aggregates and analyzes data in real time. 
A TensorFlow-based DNN is deployed in the cloud, achieving higher accuracy with larger compute resources. 
The cloud analytics layer supports model retraining, historical trend analysis, and integration with expert knowledge bases.



\subsection{Multi-Platform User Interfaces}

Farmers and agricultural experts access the system through a web, an edge, and a mobile application. 
These interfaces provide disease classification results, UAV imagery visualization, and environmental sensor data. 
When a disease is detected, the system provides actionable insights and expert-recommended solutions, allowing timely intervention.




\section{Cloud–Edge System Architecture}
\label{Sec_Architecture}

The proposed system uses Azure IoT Hub to streamline data collection and processing from multiple sources, including external IoT sensors, smartphones, edge devices, and drones, for agricultural operations (as explained in Section \ref{Sec_ProposedSystem}). 
Figure \ref{ACMS_ArchitectureDiagram} illustrates the architecture. 
This architecture separates processing into a low-latency, real-time disease-detection and a large-scale batch analytics path, while maintaining a unified data ingestion and management framework.





\subsection{Sensing Layer}

At the sensing layer, UAVs equipped with onboard cameras acquire high-resolution aerial imagery of tomato fields, focusing on visual crop assessment, which is processed locally before transmission.
Simultaneously, smartphones collect additional field imagery, while ground-based environmental sensors, such as soil moisture sensors (measure soil water content), pH sensors (measure soil acidity or alkalinity), temperature sensors (measure environmental temperature), and humidity sensors (measure air moisture), gather data specific to soil and environmental conditions. 



\subsection{IoT Edge Device Layer}

Each sensor type then preprocesses its data on respective edge devices. 
Subsequently, IoT edge devices handle buffering (temporary data storage) and inference (automated analysis or predictions), thereby ensuring resilience in rural environments with intermittent network connectivity.

\subsection{IoT Ingestion Layer}

All data streams from UAVs, mobile platforms, edge devices, and sensors are securely received through Azure IoT Hub, which handles device authentication, management, and scalable data intake. 



\subsection{Stream Analytics Routing Layer}

The incoming data is sent to Azure Stream Analytics for real-time filtering, summarization, and routing. 
At this point, the data are split into two processing paths: the real-time path, which provides immediate analysis for quick decision-making, and the batch path, which stores data for periodic, in-depth analysis.

\subsection{Real-Time Inference Layer}

The real-time path enables quick disease diagnosis and instant decision support. 
Stream Analytics sends time-critical data to Azure Functions, which drive event-based orchestration. 
These functions run deep learning models, deployed as online inference endpoints via Azure Machine Learning. 
Classification results are sent to Azure SQL Database and displayed in Power BI dashboards, enabling near-real-time crop health monitoring. 
This supports fast alerts and actionable feedback for farmers and agronomists during field operations.



\subsection{Batch Analytics Layer}

In parallel, batch processing manages long-term storage and large-scale analytics. 
Raw UAV imagery, sensor logs, and metadata are stored in Azure Blob Storage. 
Batch analytics and seasonal trend analysis run in Azure Synapse Analytics, supporting retrospective studies, model refinement, and time-series disease progression analysis. 
Azure Cognitive Search indexes data and metadata to enable efficient querying and expert reporting.



\subsection{Application Service Layer}

Outputs from the real-time and batch paths are exposed via Azure App Services, enabling secure, scalable access to system services. 
Web and mobile applications use these APIs to deliver real-time alerts, historical analytics, and recommendations, helping farmers and experts make timely, informed decisions and boost productivity.

\begin{figure*}[!t]
    \centering
    \includegraphics[width=0.51\textwidth]{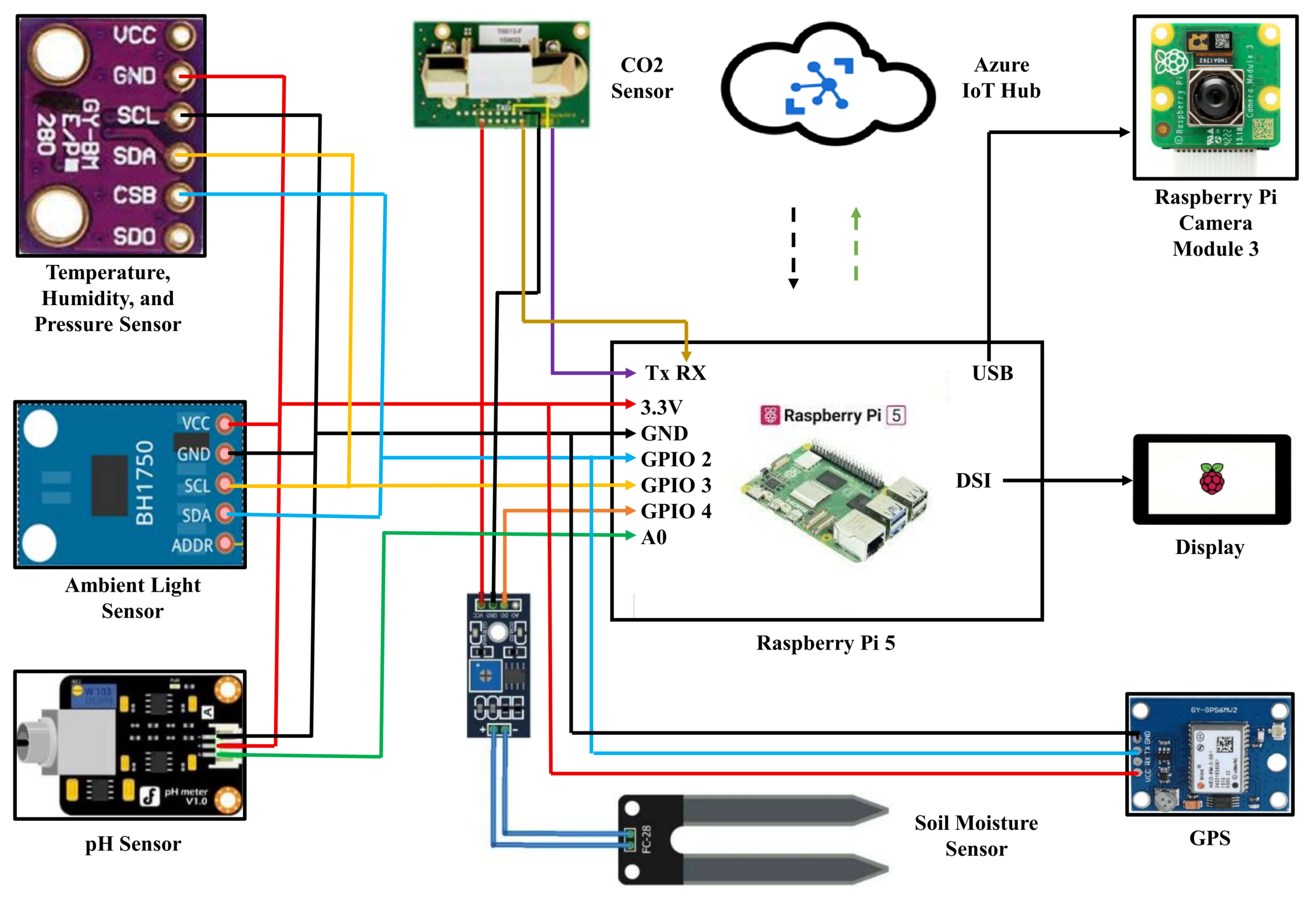}
    \caption{
    Experimental setup of the Raspberry Pi 5–based autonomous crop monitoring system, showing GPIO/I2C connections to environmental sensors (temperature–humidity–pressure, ambient light, CO2, pH, and soil moisture), GPS interfacing, Camera module 3, display integration, and cloud connectivity via Azure IoT Hub for remote monitoring and analytics.
    }
    \label{fig:ACMS_ExperimentalSetup}
\end{figure*}

\begin{figure}[!t]
    \centering
    \includegraphics[width=0.40\textwidth]{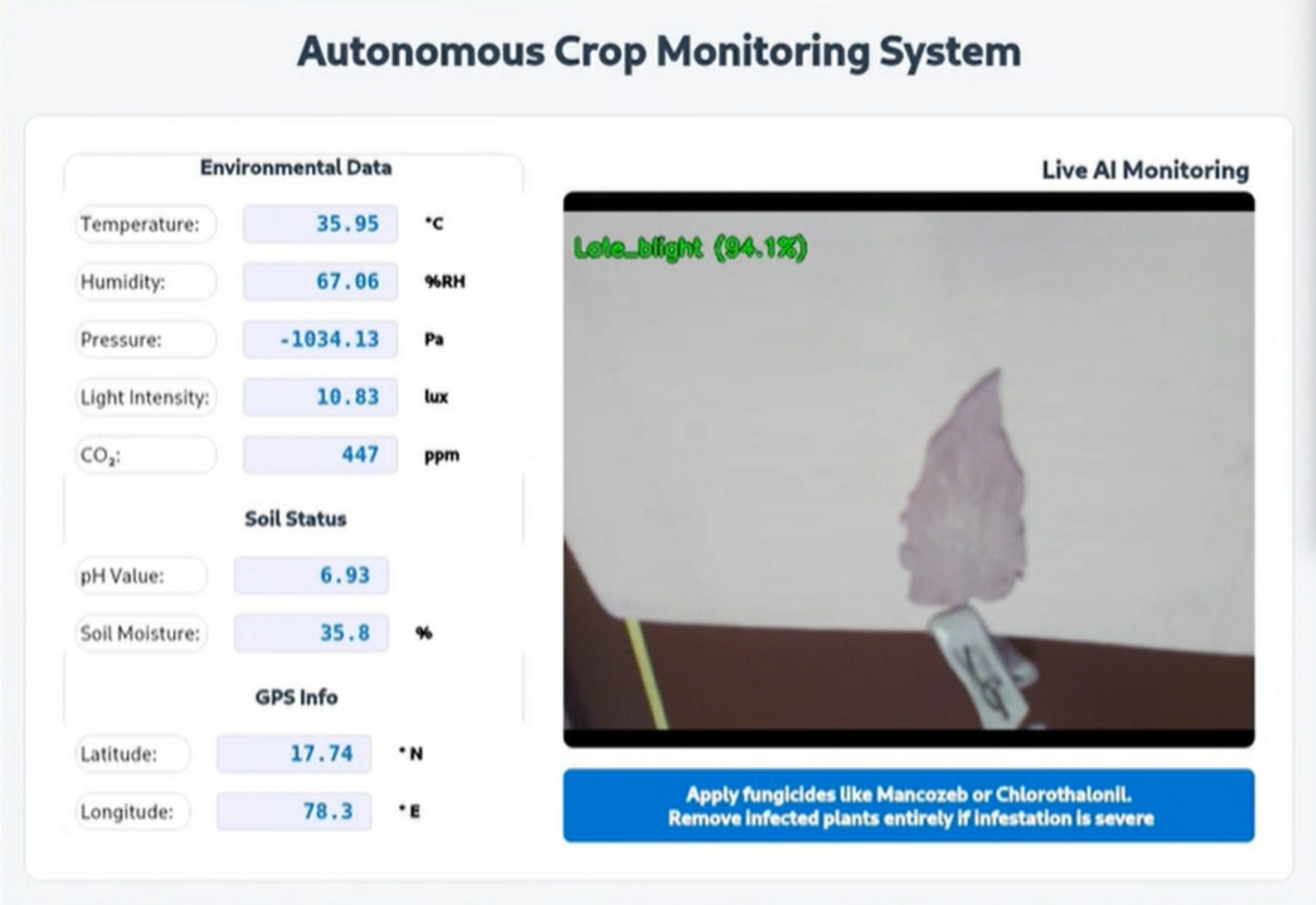}
    \caption{
    User interface results of the Raspberry Pi 5–based autonomous crop monitoring system, showing real-time sensor readings (temperature, humidity, pressure, ambient light, CO2, pH, and soil moisture), GPS information, and the captured leaf image used for on-device disease inference.
    }
    \label{fig:ACMS_ExperimentalResults}
\end{figure}

\begin{table}[!t]
\centering
\caption{
System-level performance of the proposed ACMS across mobile, web, edge device, and cloud-based deployments. 
}
\label{tab:ACMS_PerformanceEvaluation}
\begin{tabular}{lcccc}
\toprule
\textbf{Platform} & \textbf{Accuracy (\%)} & \textbf{Precision} & \textbf{Recall} & \textbf{F1-score} \\
\midrule
Mobile App   & 92 & 0.91 & 0.90 & 0.905 \\
Web App       & 95 & 0.95 & 0.94 & 0.945 \\
Raspberry Pi  & 93 & 0.92 & 0.91 & 0.915 \\
\bottomrule
\end{tabular}
\end{table}

\section{Multi-Stage System Design} \label{Sec_Design}

The design of the proposed system follows the cloud-edge system architecture as described in Section \ref{Sec_Architecture}.

\subsection{Image and Data Acquisition}

\subsubsection{UAV Imaging} 

UAVs equipped with RGB cameras were deployed to capture aerial images of tomato fields at various growth stages. 
The flight missions were planned to ensure consistent coverage and minimize blind spots.

\subsubsection{IoT Sensors}

IoT devices, including capacitive soil moisture sensors, glass electrode pH sensors, and digital light, temperature, and humidity probes, were installed in the field to collect environmental parameters. 
This multimodal data provides context for disease analysis.

\subsubsection{Dataset Preparation}

For training and validation, we used publicly available datasets, such as PlantVillage \cite{mohanty2016using} and Kaggle \cite{Sharma2018PlantDiseasesKaggle}. 

\subsection{Deep Learning Model}

\subsubsection{Model Architecture}

A TensorFlow-based DNN model was selected for disease classification, with MobileNetV2 for resource-constrained platforms (Raspberry Pi 5 or Nvidia Orin Nano Super) and ResNet-50 for cloud-based analysis.

\subsubsection{Training and Optimization}

The models were trained using TensorFlow with transfer learning from ImageNet-pretrained weights. 
The training process used categorical cross-entropy loss with the Adam optimizer, spanning over 50 epochs and a batch size of 32. 

\subsubsection{Edge Optimization}

For Raspberry Pi 5 deployment, quantization and pruning techniques were applied to reduce model size and inference time while maintaining accuracy.

\subsection{UAV-Enabled Monitoring}

UAVs autonomously followed pre-programmed flight paths over the crop field. 
The captured images were processed locally (classification on the Raspberry Pi 5) and transmitted via the Azure IoT Hub to the cloud for advanced analytics. 
This dual-mode capability ensures that disease detection remains operational in both online and offline modes.

\subsection{Azure IoT Hub Integration}

Sensor and UAV data were transmitted to Azure IoT Hub, where Stream Analytics performed real-time filtering and aggregation.
The processed data was stored in an SQL database and analyzed by TensorFlow-based AI pipeline hosted in the cloud. 
\section{Implementation} \label{Sec_Development}



The implementation of the proposed system follows the multi-stage design methodology as described in Section \ref{Sec_Design}.


\subsection{Image and Sensor Data Acquisition}

Visual data were captured using a Raspberry Pi Camera Module 3 interfaced with a Raspberry Pi 5. 
Environmental sensor data, including soil moisture, pH, light, temperature, GPS, and humidity, were acquired through GPIO and I2C-connected sensor modules.


\subsection{Deep Learning Model Implementation}

The deep learning models for tomato disease detection were developed using TensorFlow 2.13 and Keras in Python 3.10. DNN models were trained offline on a PC and optimized for deployment on edge and cloud platforms.




\subsection{Azure IoT Hub Integration}

Cloud connectivity and system orchestration were implemented using Microsoft Azure services. Azure IoT Hub serves as the centralized ingestion layer for edge device data and sensor metadata. Real-time data routing and filtering are performed using Azure Stream Analytics.




\subsection{User Interface Applications}

The mobile application was developed using Java (JDK 17) in Android Studio Giraffe (2023.3) and integrates TensorFlow Lite for on-device inference. The web application backend was implemented using Python 3.10 and the Flask framework, hosted on Azure App Service. Frontend components were developed using HTML5, CSS, and JavaScript, enabling cross-platform access through standard web browsers.

\begin{figure}[!t]
\centering
\includegraphics[width=0.40\textwidth]{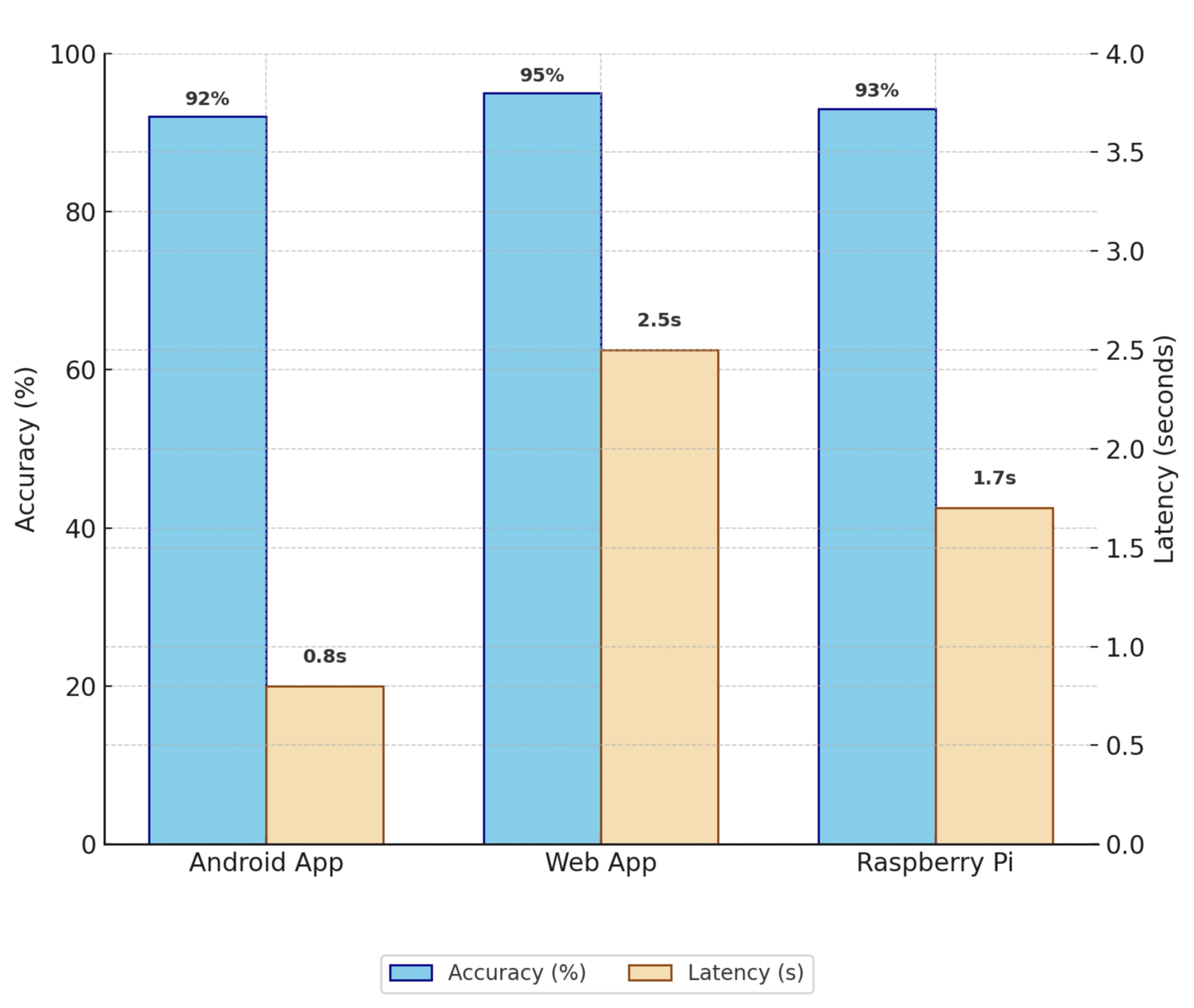}
    \caption{
    Classification accuracy and inference latency for the proposed autonomous crop monitoring system across three deployment modes: i) Mobile (on-device), ii) Web application (cloud inference via Azure IoT Hub), and iii) Raspberry Pi 5 (edge inference). 
    }
    \label{fig:ACMS_EvaluationResults}
\end{figure}

\section{Evaluation, Results, and Discussion} \label{Sec_ExperimentsAndResults}


We evaluated the proposed system in terms of sensing, inference, and application effectiveness, as described in Section \ref{Sec_Development}. 
Using the experimental setup shown in Figure \ref{fig:ACMS_ExperimentalSetup}, we assessed the system performance, as illustrated in Figure \ref{fig:ACMS_ExperimentalResults}, using standard classification metrics, including accuracy, precision, recall, and F1-score, as summarized in Table \ref{tab:ACMS_PerformanceEvaluation}.
The system achieved an overall accuracy of 92--95\%, with precision ranging from 0.91 to 0.95 and recall ranging from 0.90 to 0.94. The end-to-end latency, measured from image capture to inference and result delivery, averaged less than 8 seconds for the real-time pipeline.
Figure \ref{fig:ACMS_EvaluationResults} compares the accuracy and latency across mobile, web, and edge computing platforms, highlighting the trade-off between low-latency edge inference and higher-accuracy cloud processing. These results demonstrate that integrating on-site sensing with edge- and cloud-based AI analytics enables accurate and scalable crop disease monitoring.

\section{Conclusion} \label{Sec_Conclusions}

This paper presents an autonomous crop-monitoring system that integrates IoT sensing, edge intelligence, and cloud analytics to enable real-time, large-scale detection of crop diseases. 
The proposed system was designed as a cloud–edge architecture with explicitly separated real-time inference and batch analytics paths, enabling both low-latency decision support and long-term disease trend analysis.
By combining deep–learning–based visual diagnosis with IoT-driven data ingestion and scalable cloud services, the system addresses key limitations of manual crop inspection and conventional vision-based disease-detection approaches.
Experimental results indicate consistently high classification performance across mobile, web, and edge deployments, with an average end-to-end latency below 8 seconds. 
The modular design and standardized interfaces also enable the incorporation of additional IoT sensors, UAV platforms, and trained models with minimal reconfiguration, improving accessibility for both farmers and agricultural experts. 
Future work will focus on extended field trials, robustness to real-world variability, and tighter integration with autonomous UAV and decision-making to enable adaptive flight planning, navigation and closed-loop crop monitoring for sustainable agriculture.

\addtolength{\textheight}{-12cm}   






\bibliographystyle{IEEEtran}
\bibliography{References}

\end{document}